\documentclass[sigconf]{acmart}
\AtBeginDocument{%
  }

\copyrightyear{2025}
\acmYear{2025}
\setcopyright{acmlicensed}\acmConference[WWW '25]{Proceedings of the ACM Web Conference 2025}{April 28-May 2, 2025}{Sydney, NSW, Australia}
\acmBooktitle{Proceedings of the ACM Web Conference 2025 (WWW '25), April 28-May 2, 2025, Sydney, NSW, Australia}
\acmDOI{10.1145/3696410.3714778}
\acmISBN{979-8-4007-1274-6/25/04}

\usepackage{amsmath}
\usepackage{amssymb}
\usepackage{booktabs}
\usepackage{soul}
\usepackage{multirow}
\usepackage{colortbl}
\usepackage{subfig}
\usepackage{graphicx}

\usepackage{algorithm}
\usepackage{algorithmic}
\usepackage{latexsym}
\usepackage{microtype}
\usepackage{graphics}
\usepackage{arydshln}
\usepackage{diagbox}
\usepackage{float}
\usepackage{enumitem}
\usepackage{bbding}
\usepackage{array}
\usepackage{appendix}
\usepackage{times}
\usepackage{caption}
\usepackage{xcolor}
\usepackage{newfloat}
\usepackage{listings}
\usepackage{pifont}
\usepackage{subcaption}  
\usepackage{tcolorbox}
\usepackage{anyfontsize}
\usepackage[T1]{fontenc}
\usepackage[utf8]{inputenc}

\newcommand{\model}{\texttt{M3H}}
\newcommand{\data}{\texttt{AxiOM}}


\begin{document}
\title{Figurative-cum-Commonsense Knowledge Infusion for Multimodal Mental Health Meme Classification}

\author{Abdullah Mazhar}
\affiliation{%
  \institution{IIIT Delhi}
  \city{New Delhi}
  \country{India}}
\email{abdullahm@iiitd.ac.in}

\author{Zuhair Hasan Shaik}
\affiliation{%
  \institution{IIIT Dharwad}
  \city{Dharwad}
  \country{India}}
\email{zuhashaik12@gmail.com}

\author{Aseem Srivastava}
\affiliation{%
  \institution{IIIT Delhi}
  \city{New Delhi}
  \country{India}}
\email{aseems@iiitd.ac.in}

\author{Polly Ruhnke}
\affiliation{%
  \institution{University of Illinois Chicago}
  \city{Chicago}
  \country{USA}}
\email{pruhn2@uic.edu}

\author{Lavanya Vaddavalli}
\affiliation{%
  \institution{University of Illinois Chicago}
  \city{Chicago}
  \country{USA}}
\email{lvadd@uic.edu}

\author{Sri Keshav Katragadda}
\affiliation{%
  \institution{University of Illinois Chicago}
  \city{Chicago}
  \country{USA}}
\email{skatra5@uic.edu}

\author{Shweta Yadav}
\affiliation{%
  \institution{University of Illinois Chicago}
  \city{Chicago}
  \country{USA}}
\email{shwetay@uic.edu}

\author{Md Shad Akhtar}
\affiliation{%
  \institution{IIIT Delhi}
  \city{New Delhi}
  \country{India}}
\email{shad.akhtar@iiitd.ac.in}

\renewcommand{\shortauthors}{Mazhar et al.}

\begin{abstract}
The expression of mental health symptoms through non-traditional means, such as memes, has gained remarkable attention over the past few years, with users often highlighting their mental health struggles through figurative intricacies within memes. While humans rely on commonsense knowledge to interpret these complex expressions, current Multimodal Language Models (MLMs) struggle to capture these figurative aspects inherent in memes. To address this gap, we introduce a novel dataset, \data, derived from the GAD anxiety questionnaire, which categorizes memes into six fine-grained anxiety symptoms. Next, we propose a commonsense and domain-enriched framework, \model, to enhance MLMs’ ability to interpret figurative language and commonsense knowledge. The overarching goal remains to first understand and then classify the mental health symptoms expressed in memes. We benchmark \model\ against $6$ competitive baselines (with $20$ variations), demonstrating improvements in both quantitative and qualitative metrics, including a detailed human evaluation. We observe a clear improvement of $4.20\%$ and $4.66\%$ on weighted-F1 metric. To assess the generalizability, we perform extensive experiments on a public dataset, \texttt{RESTORE}, for depressive symptom identification, presenting an extensive ablation study that highlights the contribution of each module in both datasets. Our findings reveal limitations in existing models and the advantage of employing commonsense to enhance figurative understanding.

\end{abstract}
\settopmatter{printfolios=true}
\maketitle

\begin{figure}[t]
    \centering
    \includegraphics[width=\columnwidth]{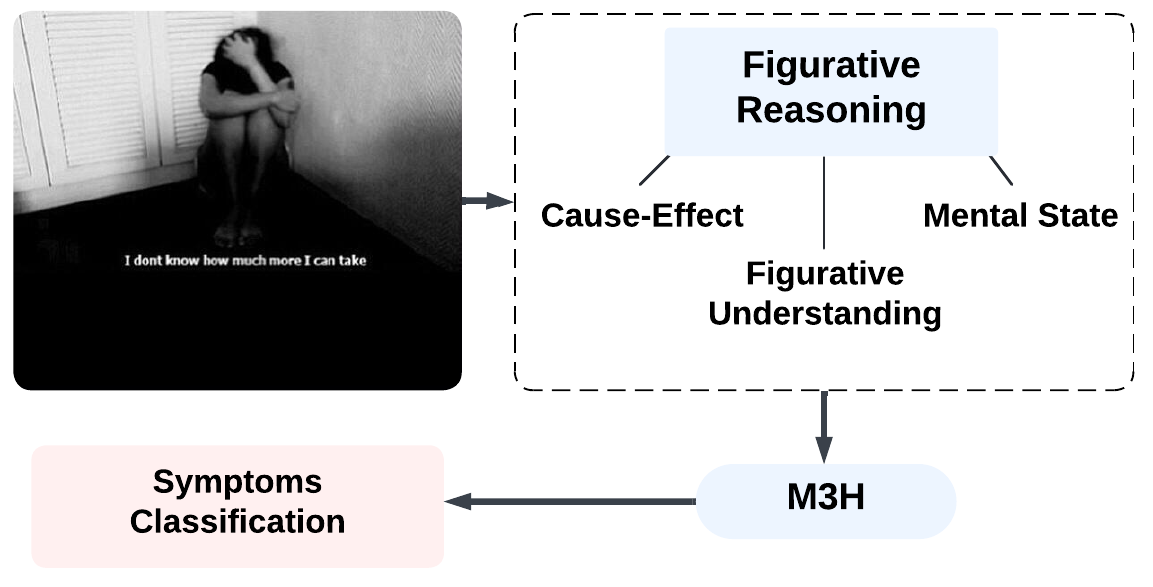}
    \caption{Our problem statement involves the classification of mental health memes focusing on depressive and anxious social media content. We employ figurative reasoning in our proposed method, \model, for symptom classification.}
    \label{fig:example}
    \vspace{-7.5mm}
\end{figure}

\section{Introduction}
In recent years, the increased use of social media platforms such as Reddit for sharing memes has emerged as a fascinating form of communication, especially among individuals expressing mental health symptoms \cite{akram2022mental}. These seemingly light-hearted memes often carry deep, hidden meanings that reflect complex emotional states. For many in contemporary society, memes are not just a source of humor but a creative outlet to subtly convey thoughts and feelings that are difficult to articulate directly. Yet, the true essence of these memes, especially those with mental health nuances, remains elusive to many. This happens primarily due to the inherent figurative language they often contain. To truly comprehend the core meaning behind these memes, humans require two major fronts: (a) commonsense knowledge and (b) figurative understanding. Together, these elements decode the deeper meaning embedded within a meme, revealing not only its humor but also the domain context behind it. For example, as shown in Figure \ref{fig:example}, the person is sitting in a corner, holding head in distress, which suggests deep emotional pain or hopelessness. The text, {\em ``I don't know how much more I can take"}, hints at overwhelming mental strain, possibly even suicidal thoughts. Here, commonsense understanding helps us interpret the person’s body language and posture as signs of despair. Also, the figurative understanding is captured by the darkness of the image, symbolizing the person's inner turmoil and dark thoughts they may be experiencing. However, this process of interpretation, while relatively straightforward for human cognition, becomes significantly more complex when applied to artificial intelligence (AI).

Current advancements in visual language models (VLMs) and large language models (LLMs) continue to struggle with understanding sarcasm, satire, and nuanced figurative expressions in memes, especially in the context of mental health. These models have difficulty capturing the layered meanings that are essential for understanding such content \cite{Zhong_2024_CVPR}. Other works in this space have explored various facets of meme research. For instance, \citet{electronics13142780} developed a hateful meme detection method, focusing on the complexity of weakly related visual and textual modalities in memes. Other works show enhancement in meme detection by leveraging global and local perspectives and addressing harmful memes through multimodal analysis \cite{pramanick-etal-2021-momenta-multimodal}. \citet{sharma2022detecting} further surveyed the complexities of harmful meme detection, highlighting the gaps in datasets, particularly for under-explored topics like self-harm and extremism. However, we barely find research dedicated to the study of memes for sensitive domains like mental health. On the other hand, the primary cause of limited research in this space is possibly due to the limited datasets \cite{wsdmAseem}. To the best of our knowledge, we find \texttt{RESTORE} to be the only publicly available dataset for depressive meme classification \cite{yadav2023towards}. To address this gap, we first propose a novel dataset, \data, containing memes derived from the GAD anxiety questionnaire. We choose this issue to introduce diversity in the existing set of data. \data\ consists a total of $3582$ memes. We employ annotators to categorize each meme into anxiety symptoms. Healthcare professionals further validate the efficacy of our annotation guidelines and the quality of the dataset produced. Next, we propose a novel multimodal framework, \model, to classify mental health symptoms associated with each meme. We focus on employing figurative understanding with commonsense reasoning using VLMs. The core of our framework lies in utilizing these VLM's capabilities and further enhancing them through the dedicated retrieval-augmented-generation (RAG) method, allowing us to navigate the complex humor and underlying emotions within memes through knowledge fusion. For instance, as shown in Figure \ref{fig:example}, our method first attempts to fetch figurative  reasoning through commonsense with three major attributes: (a) cause-effect, (b) figurative-understanding, and (c)  mental-state. Next, a classifier is responsible for symptom classification.

In our work, \model\ contains GPT-4o for figurative reasoning and a BART-based transformer that allows the final encoding and classification of symptoms. We rigorously benchmark \model\ against 6 competitive baseline methods (with 20 variations), including the current state-of-the-art system, on both the proposed \data\ and \texttt{RESTORE} dataset. We employ standard classification metrics such as macro-F1 and weighted-F1 scores to evaluate performance, revealing significant improvements of $4.94\%$ and $5.79\%$ on macro-F1; whereas an improvement of $4.20\%$ and $4.66\%$ on weighted-F1 across \data\ and \texttt{RESTORE}, respectively. Additionally, we present an exhaustive human evaluation and qualitative and quantitative analysis to further understand where our method excels and what it lacks. Our findings state the efficacy of our approach in bridging the gap between human-level meme comprehension and AI's current capabilities. Our contributions are summarized below:

\begin{itemize}[leftmargin=*]
    \item \textbf{Novel Dataset}: We propose a novel dataset, \data, derived from the GAD anxiety questionnaire, which categorizes memes into six fine-grained anxiety symptoms. 
    \item \textbf{Novel Framework}:We propose a novel framework, \model, that employs figurative-cum-commonsense understanding for enhanced interpretation of memes related to mental health subjects and further classify associated symptoms.
    
    \item \textbf{Benchmarking}: We compare the performance of \model\ against 6 baselines (with 20 variations) and provide extensive ablation studies and human evaluations to test the efficacy of our method. 
    \item \textbf{Generalizability}: To validate the generalizability of our proposed model, we benchmark it on an additional dataset, \texttt{RESTORE}, where \model\ consistently outperforms baseline methods. 
\end{itemize}

\noindent{\bf Reproducibility.} The code and dataset (for research purposes only) {are available at} \url{https://github.com/flamenlp/M3H}.

\section{Related Work}

Our work builds upon three key areas of prior research: the study of memes and social media, mental health expressions in online spaces, and multimodal approaches to understanding mental health content. Here, we review the most relevant studies in each of these domains.

\paragraph{\bf Social Media and Memes.} Memes have become a widely-used medium for expressing ideas and communicating on social media platforms. Researchers have developed techniques to detect and categorize memes within large datasets of social media content, leveraging the meme’s visual and textual elements, metadata, and patterns of diffusion across networks \cite{10.1007/978-3-319-71767-8_87, suryawanshi-chakravarthi-2021-findings, pasha2024multilingual}. Other studies have focused on the cross-platform spread of memes, providing valuable insights into how they are shared and reused across various social media environments \cite{joshi2024contextualizing}.
The phenomenon of memes going "viral" has also been extensively examined, with researchers discovering that even simple images or phrases can rapidly gain traction online, forming temporary communities centered around their shared content \cite{sanford2023going,
taecharungroj2014effect,
md2020identifying}. In addition, some studies have specifically analyzed video memes, exploring how short video clips are repeatedly remixed and repurposed, often as a means of disseminating news \cite{xie2011visual,laskin2022memes,du2020understanding}.
Memes have also proven to be a valuable tool in marketing. Research indicates that humorous memes tend to be more effective than serious imagery in social media marketing campaigns, particularly when they garner high audience engagement \cite{malodia2022meme,vasile2021using}.

\paragraph{\bf Social Media and Mental Health.}
The impact of social media on mental health, especially for young people, is an important area of study. One study found that when Facebook was introduced at colleges, it had a negative effect on students' mental health \cite{esco2023perceived,
jones2022perceptions}. However, other research on young adults has shown mixed results. Some studies suggest social media might harm mental health, while others find no evidence of harm or even some benefits \cite{akram2022mental,gupta2021exploring,kariko2019laughing,myrick2022consuming, aseemPlosOne}.
Researchers have also looked at ways to predict mental health issues using social media data. However, a review of these methods found that there's a need for better standards and more valid ways of measuring mental health in these studies \cite{yadav2023towards,boinepelli2022towards,hamza2024multimodal,pramanick2021momenta}.
During the COVID-19 pandemic, researchers found that people who used social media a lot were more likely to have mental health problems \cite{gupta2021exploring}.

\paragraph{\bf Multimodal Approaches and Commonsense Reasoning in Mental Health.} 
Recent research has started to use information from multiple sources to understand mental health better. Previous studies used information from text, images, and how people interact on social media to analyze mental health \cite{de2013role, braghieri2022social, aseemICWSM}. Some researchers are using data from social media activities and physical signals (like heart rate) to monitor mental health without being intrusive \cite{zhou2015tackling, hickey2021smart}. This approach could help detect mental health issues early. A new method has been developed to detect emotions and emotional reasoning in conversations \cite{bhaumik2024towards}. It uses information from different sources and applies commonsense reasoning. This is particularly relevant to our work, as understanding memes often requires similar skills. Recently, researchers have proposed a new way to classify mental health using multiple types of data and a technique called ``knowledge distillation'' \cite{khan2021novel}. Additionally, \citet{srivastava-etal-2024-knowledge} demonstrated the effectiveness of LLMs in mental health tasks like counseling summarization, further highlighting the potential of structured knowledge alignment. Our work builds on these studies by combining the analysis of memes, the focus on mental health in online spaces, and the use of multiple types of data and commonsense reasoning. We aim to create a more complete understanding of how depressive symptoms are expressed through memes.

\paragraph{\bf Depression Analysis.} The RESTORE dataset has been instrumental in advancing depression analysis through memes. This dataset is unique in its annotation of fine-grained depression symptoms based on the clinically adopted PHQ-9 questionnaire \cite{manea2015diagnostic}.
We perform sota work upon this dataset, and \textbf{our findings mention that we need to consider both visual and textual commonsense together at once. Traditional multimodal approaches that encode visual and textual information separately before fusing them may miss crucial implicit connections between these elements.}

\begin{figure*}[ht]
    \centering
        \includegraphics[width=\textwidth]{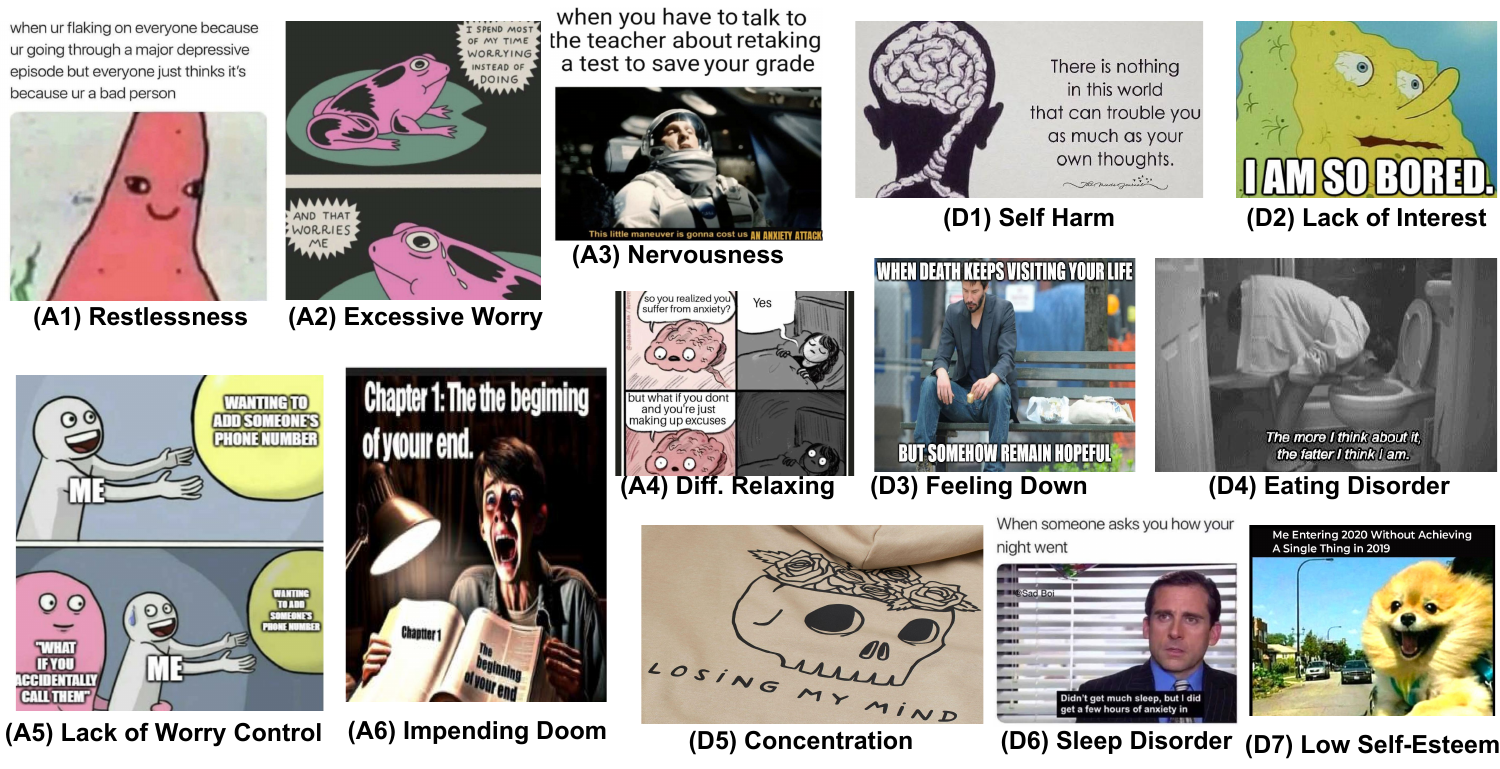}
    \caption{Label-wise sample of memes in the proposed dataset, \data\ (A1 -- A6), illustrating different GAD anxious symptoms and \texttt{RESTORE} (D1 -- D7), illustrating different PHQ-9 depressive symptoms.}
    \label{fig:dataset_samples}
    \vspace{-3mm}
\end{figure*}

\section{Dataset}
In this section, we discuss the dataset construction pipeline and the measures taken to ensure the quality of annotations followed by analysis. We gather data from various online sources, including Reddit and Instagram. We discuss the standard collection process and standards in Appendix (c.f Section \ref{dataCollection}).

\subsection{Annotation Guidelines}
We utilize the {\em Labelbox} interface to annotate the memes based on the GAD questionnaire, which assesses anxiety severity. With the help of experts, we annotate the memes into six categories, each representing a symptom of anxiety. 
We establish the annotation guidelines in consultation with healthcare professionals and individuals diagnosed with anxiety. These definitions helped our annotators categorize memes according to the specific GAD symptoms. Below, we describe the definitions of each label.

\begin{itemize}[leftmargin=*]
    \item \textbf{Nervousness:} This category represents mild anxiety, typically related to common, short-term stressors. Memes in this category depict minor worries that cause nervousness but are manageable and not excessive. 
    \begin{itemize}
        \item {Indications:} This GAD category carries an indication of a person rubbing their forehead, temples, or hands. The text talks about something minor that made a person nervous but clearly won’t last long, isn’t excessive, and is manageable.
    
        \item {Example:} A meme featuring a scared-looking man with the caption {\em "when you have to talk to the teacher about retaking a test to save your grade. This little maneuver is gonna cost us AN ANXIETY ATTACK"} showcases minor, temporary nervousness.
    \end{itemize}

    \item \textbf{Lack of Worry Control:} This refers to anxiety that appears out of nowhere or feels intrusive, often overwhelming the individual. Memes in this category often reflect uncontrollable worry or the use of unhealthy coping mechanisms, such as substance use.
    \begin{itemize}
        \item Indications: This sub-category is entailed by a visual of something or someone jumping into the picture or lurking in the background. The text is along the lines of {\em "My anxiety:"} to show the lack of control over their anxiety.
    
        \item {Example:} A meme with the text {\em ``Me: Wanting to add someone's phone number"} paired with an image of a worried character being stopped by an anxious figure saying {\em ``What if you accidentally call them?"} implies a sudden, uncontrollable spike. 
    \end{itemize}
    
    \item \textbf{Excessive Worry:} This category involves persistent, excessive worry about both trivial and significant matters. Memes in this category typically feature visuals of profuse sweating or expressions of extreme stress.
    \begin{itemize}
        \item {Indications:} This GAD category is entailed by the visuals having excessive sweating or shaking. The text expresses stress.

        \item {Example:} A meme featuring a pink frog sitting on a lily pad with the text {\em ``I spend most of my time worrying instead of doing"} and {\em ``And that worries me"} reflects a sense of self-anxiety.
    \end{itemize}
    
    \item \textbf{Difficulty Relaxing:} Memes in this category depict individuals struggling to rest or relax, often due to racing thoughts. These memes may feature visuals of someone unable to sleep or unwind despite being in a relaxing setting.
    \begin{itemize}
        \item {Indications:} This GAD category may be entailed by the visuals of a person in a relaxing setting, like a bed or beach, with eyes wide. It is usually accompanied by text about overthinking, but because they can’t seem to relax, it falls into the category of difficulty relaxing instead of excessive worry. Sometimes, the image can be of someone staring off into space with the text saying something along the lines of "Me at five AM:". This means they cannot rest and, therefore, are categorized as having difficulty relaxing.

        \item {Example:} A meme featuring a four-panel meme showing a brain talking to a person in bed. In the first panel, the brain says, {\em "So you realized you suffer from anxiety?"} to which the person responds {\em "Yes"} while trying to sleep. In the next panel, the brain follows up with, {\em "But what if you don't and you're just making up excuses?"} causing the person to be wide awake in the final panel, their eyes bulging with worry.
    \end{itemize}

    \item \textbf{Restlessness:} This category includes memes that depict physical or mental discomfort, where individuals are unable to sit still or concentrate. Visuals often show characters feeling unsatisfied.
    \begin{itemize}
        \item {Indications:} This sub-category is often quite clear, as the meme often contains visuals of someone freaking out, sobbing, or tearing their hair out. The image is usually purposefully blurred or distorted, and the text usually talks about something terrible the person creating the meme had gone through.

        \item {Example:} A meme featuring a close-up of Patrick Star with the text {\em "when ur flaking on everyone because ur going through a major depressive episode but everyone just thinks it's because ur a bad person"} reflects the internal struggle of depression and the external misunderstanding by others.
    \end{itemize}
    
    \item \textbf{Impending Doom:} This category reflects feelings of looming catastrophe or fear, despite no immediate danger. Memes often depict paranoia or anxiety about something bad happening.
    \begin{itemize}
        \item {Indications:} This GAD category may be entailed by a looming presence over someone or someone panicking/worried with text that implies that the person thinks something bad is about to happen in their lives.
    
    \item {Example:} A meme showing "When you start reading a new book and realize it's predicting your doom" - depicting a terrified person screaming while reading a sinister book titled "The beginning of your end" conveys a sense of impending doom.
    \end{itemize}

\end{itemize}

\paragraph{\bf \texttt{RESTORE} Dataset.} We use an additional public meme depressive symptom classification dataset, \texttt{RESTORE} \cite{yadav-etal-2023-towards}. The authors carefully compile the dataset using a clinically-informed methodology, drawing from two widely-used social media platforms: {\em X} and {\em Reddit}\footnote{Reddit: \url{https://www.reddit.com/} ; X (formerly Twitter): \url{https://x.com/}}. The original \texttt{RESTORE} categorizes depressive symptoms across eight distinct categories based on the PHQ-9 questionnaire: {\em Feeling Down, Lack of Interest, Eating Disorder, Sleeping Disorder, Concentration Problem, Lack of Energy, Low Self-Esteem}, and {\em Self-Harm}. However, upon reviewing the dataset, we observe significant inconsistencies in the annotations associated with the {`Lack of Energy’} category. To ensure the integrity of our analysis, we exclude this label from our study, resulting in a focus on the remaining seven categories.

\subsection{Data Analysis}
\data\ consists of 3,582 memes categorized into six anxiety-related GAD symptoms, divided into training, validation, and testing sets in a 70:10:20 ratio. Healthcare professionals validated the dataset, making minor adjustments to ensure accurate categorization. Table \ref{tab:data_distribution} provides a detailed label-wise analysis, showing the distribution of each anxiety symptom across the splits. The training set, with 2,153 memes, forms the majority, providing a robust foundation for model training. The validation and test sets, containing 307 and 615 memes respectively, ensure balanced fine-tuning and evaluation. This division ensures diverse and consistent representation of anxiety expressions across all GAD categories.

Additionally, we perform a visual analysis to study the relationship between anxiety symptoms and color features, focusing on saturation (vibrancy) and value (brightness). Most memes, regardless of category, are colorful, reflecting heightened and chaotic emotions. Restlessness memes exhibit the highest vibrancy with low hue (red tones), while impending doom and difficulty relaxing memes show the least vibrancy, often featuring grayscale tones, shadows, or evening settings.

\begin{table*}[ht]
    \centering
\subfloat[Depression: RESTORE dataset]{
    \begin{tabular}{lcccccccc}
        \toprule
	          & LOI	  &	FD    &	ED	  &	SD 	  &	LSE	  &	CP	  &	SH	  &	Total \\ \midrule
        Train   & 441	&	1555	&	1714	&	997	&	549	&	348	&	1259	&	6863\\
        Val     &	13	&	55	&	35	&	18	&	29	&	20	&	27	&197\\
        Test    &	35	&	82	&	61	&	48	&	32	&	50	&	28	&336\\ \midrule
        Total   & 	489	&	1692	&	1810	&	1063	&	610	&	418	&	1314	&	7396\\
        \bottomrule
    \end{tabular}}
    \hfill
    \subfloat[Anxiety: \data\ dataset]{
    \begin{tabular}{lcccccccc}
        \toprule        
                & NV	&	LWC	&	EW	&	DR	&	RST	&	ID	&	Total \\ \midrule
        Train   &	373	&	331	&	322	&	356	&	405	&		366	&	2153 \\
        Val     &	53	&	47	&	46	&	51	&	58		&	52	&	307	\\
        Test    &	106	&	94	&	92	&	102	&	116	&		105	&	615	 \\ \midrule
        Total   &	532	&	472	&	460	&	509	&	579	&		523	&	3582	\\
        \bottomrule
    \end{tabular}
    }
    \caption{Data statistics for the depression and anxiety datasets. 
    (a) Depression Labels: LOI: \textit{Lack of interest}; FD: \textit{Feeling down}; ED: \textit{Eating Disorder}; SD: \textit{Sleeping Disorder}; LSE: \textit{Low Self-Esteem}; CP: \textit{Concentration Problem}; SH: \textit{Self Harm}
    (b) Anxiety Labels: NV: \textit{Nervousness}; LWC: \textit{Lack of Worry Control}; EW: \textit{Excessive Worry}; DR: \textit{Difficulty Relaxing}; RST: \textit{Restlessness}; ID: \textit{Impending Doom}}
    \label{tab:data_distribution}
\end{table*}

\paragraph{\bf Annotator Details, Quality, and Data Ethics.} Both datasets, \data\ and \texttt{RESTORE}, address sensitive mental health topics such as anxiety and depression, and we prioritize ethical considerations at every stage. The annotation process has been conducted by a set of six annotators. Additionally, healthcare professionals verify the quality of each iteration. The annotators represented different age groups, ranging from 25 to 38, and came from diverse cultural and educational backgrounds, ensuring a broad perspective on mental health symptoms. We source data from publicly available platforms, and all personally identifiable information (PII) is thoroughly anonymized to protect user privacy. Furthermore, we calculate an agreement score of $81\%$ using the Intersection Union Agreement ($IUA$) method \cite{choudhary2005assessment}, which falls under the strong annotator agreement category.

\section{Methodology}
In this section, we discuss the complete pipeline of our proposed framework, \model, and its contributory modules.

\begin{figure*}[ht]
    \centering
    \includegraphics[width=\textwidth]{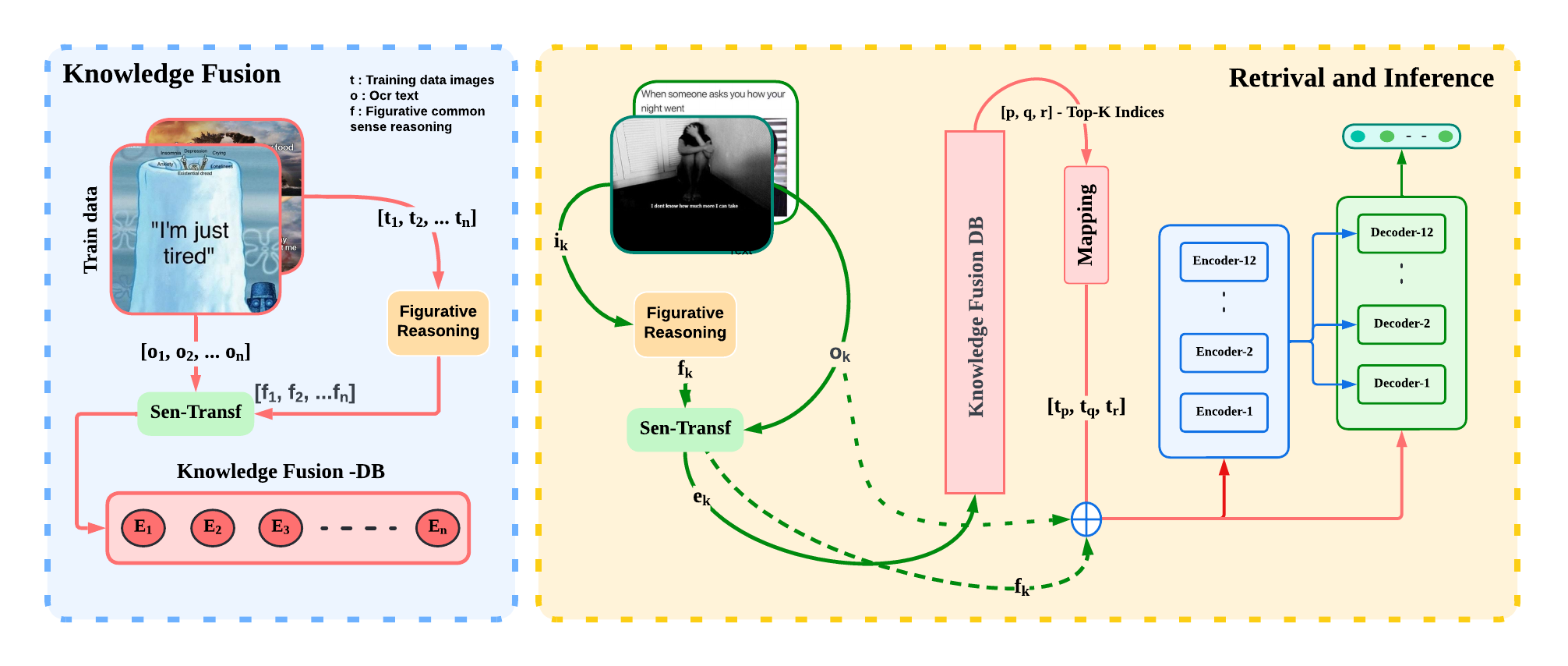}
    \caption{Proposed Framework: \model. The meme image and OCR text are standalone inputs to our framework. Primarily, we employ LLM for figurative reasoning on prominent commonsense attributes. We deploy the reasoning into a RAG-database to be considered for knowledge fusion during classification. Later, we employ an encoder-decoder framework for final classification.}
    \label{fig:arch}
    \label{-6mm}
\end{figure*}

\paragraph{\bf Problem Statement.} We address a standard multi-modal image-cum-text classification task. Given a meme image ($I$) from social media consisting of OCR-extracted text ($O$), the objective is to classify associated symptom ($s_i$). To achieve this, we fetch and utilize (a) figurative semantics and (b) exemplar knowledge. In the following sections, we detail how to retrieve these insights from memes followed by its integration into \model\ for classification. 

\subsection{Figurative Reasoning}
Memes frequently convey layered meanings that are challenging to interpret, even for humans, as they require a deep understanding of commonsense knowledge to figure out the underlying meaning of memes. This knowledge, {\em aka} figurative knowledge, inherits information related to causality, figurative expression, and cognitive understanding. To address this complexity, we employ prompt engineering on state-of-the-art LLMs to generate commonsense-enriched, figurative reasonings for each meme. We generate these reasonings based on three key attributes that capture different aspects of meme interpretation:

\begin{itemize}[leftmargin=*]
    \item \textbf{Cause-Effect:} Memes often depict scenarios that reflect real-world experiences or situations with clear causes and outcomes. Identifying these cause-effect relationships helps to ground the meme's content in reality, allowing the model to better relate the visual and textual elements to real-life consequences, which are vital for understanding expressions such as anxiety or stress.
    
    \item \textbf{Figurative Understanding:} Memes are rich in figurative language, often using metaphors, analogies, or symbolic representations to convey deeper, sometimes hidden, messages. Recognizing these figurative elements is essential for decoding the true intent of the meme, which may be cloaked in humor, irony, or satire-common ways users express their  struggles indirectly.

    \item \textbf{Mental State:} It is about recognizing the specific psychological state being expressed in the meme.
\end{itemize}

We employ prompt to fetch this figurative reasoning ($r$) in order to assist the framework's pipeline in better understanding the meaning and nuances of humor underlying the meme. We present prompt and sample reasoning text in Appendix (c.f Section \ref{promptsandexamples}).

\subsection{Knowledge Fusion}
To overcome the limitation of static models being unable to take advantage of dynamic artifacts, we fuse dedicated knowledge into the \model through a Retrieval-Augmented Generation (RAG) module. This method allows for more contextual and up-to-date meme comprehension and ensures that \model\ is informed by the most relevant and recent information. We utilize images ($I$) in the train split with their figurative reasoning ($r$) and OCR text ($o$), $I_t \in \{(o_1,r_1) \dots (o_n,r_n)\}$ to yield most relevant instances from the embedded knowledge, as mentioned below in Equation \ref{eq:rag}.
\begin{equation}
\mathcal{E} =  \Pi([o_1, o_2, \dots, o_n])^{{n \times d}} \oplus \Pi([r_1, r_2, \dots, r_n])^{{n \times d}}
\label{eq:rag}
\end{equation}
where $\Pi$ is a sentence-transformer that yields  linear embeddings corresponding to each tuple $\langle o_x, r_x \rangle$. We create the RAG database using the concatenated embeddings of both OCR text and reasoning, resulting in a database of final RAG-Embedding, $\mathcal{E} \in \mathbb{R}^{n \times 2d}$.

For each new meme as an input, we retrieve the top-\(n\) most relevant meme indices, $\gamma$,
from the RAG database. To achieve this, we compute the embedding for the input meme by concatenating the individual embeddings of OCR-text \( o_k \) and the figurative content \( r_k \), as shown below.
\begin{equation}
    e_k = (\Pi(o_k) \oplus \Pi(r_k)) \in \mathbb{R}^{1 \times 2d} ; \gamma = \arg \max_{i} \Phi(e_k,\mathcal{E}_i)
\end{equation}
The embedding \( e_k \) is then compared against the database, $\mathcal{E}$ using cosine similarity $\Phi$, and the output consists of the indices of top-\(n\) similar instances, which correspond to the \(n\) most similar memes.

\subsection{Classifier}
Once the relevant examples are retrieved, we combine all the information to construct input for our classifier. The input is designed to guide the classifier in accurately classifying mental health issues depicted in the memes. Our method enhances the understanding of memes through knowledge fusion, and the final classifier leverages text modality to classify symptoms. We present complete prompt input with a few examples in the Appendix (c.f. Section \ref{promptsandexamples}, Table \ref{promptInputExample}).  Precisely, we inculcate the following attributes in our final input:  

\begin{itemize}[leftmargin=*]
    \item \textbf{OCR-Text} ($o$): The text extracted from the meme forms an objective component of the classifier’s input. This text often includes literal content, such as captions or dialogue within the meme.
    \item \textbf{Figurative Reasoning} ($r$): Figurative language, as they often use humor, irony, and metaphors to communicate complex states. It generally can be comprehended based on commonsense. 
    \item \textbf{Relevant Knowledge Attributes} ($\gamma$): It involves retrieving relevant information from the RAG module (knowledge-fusion). 
\end{itemize}

As a result, we employ an encoder-decoder framework that takes the combined input and classifies associated symptoms. This few-shot learning approach, supported by the retrieval of similar examples, enhances the model's ability to generalize from limited data, making it well-suited for analyzing the nuanced nature of memes.

\paragraph{\bf Training}
We train the entire framework for two different datasets, namely (a) \data\ and (b) RESTORE. Here, the primary dataset possesses a multiclass symptom classification problem, whereas the latter is a multilabel setting. For \model\ to accommodate this requirement, we put a classifier personalized to each of these problem statements. For \data, we put a single softmax for a single class. On the other hand, for RESTORE, the classification for each class \( y_j \) is determined as follows:
\begin{equation}
y_j = 
\begin{cases} 
1 & \text{if } P(c_j \mid o_k, r_k, \gamma_k) > \tau_j, \\
0 & \text{if } P(c_j \mid o_k, r_k, \gamma_k) \leq \tau_j,
\end{cases}
\end{equation}
where $\gamma_k$ represent the set of top-\(n\) similar examples, and we set $\tau_j$ as threshold for multilabels.

\section{Experiments, Results, and Analysis}
Here, we discuss the selection of baselines and compare their performance with the proposed method, \model. Further, we present an exhaustive human evaluation and analysis.

\subsection{Baselines}
In our work, we experiment with the following LLMs for figurative reasoning:  
$\blacktriangleright$\textbf{COMET} \cite{bosselut2019cometcommonsensetransformersautomatic}: Commonsense transformer with better contextual understanding of textual information by capturing commonsense relationships. $\blacktriangleright$\textbf{LLAVA} \cite{liu2023visualinstructiontuning}: It is a multimodal LLM pretrained on textual and visual data providing better cross-modal modeling. In our case, it is observed to understand complex figurative nuances well. $\blacktriangleright$\textbf{GPT-4o} \cite{openai2024gpt4technicalreport}: GPT-4o is state-of-the-art LLM capable of handling both textual and visual data. It performs best due to its ability to interpret multimodal and diverse nuances.
Next, we experiment with the following classification heads within our framework:
$\blacktriangleright${\bf BERT} \cite{devlin2019bertpretrainingdeepbidirectional}: A widely known encoder in NLP literature, often compared with other models like DeBERTa.
$\blacktriangleright${\bf DeBERTa} \cite{he2021debertadecodingenhancedbertdisentangled}: A notable extension of BERT, incorporating disentangled attention and more advanced techniques for improved contextual understanding.
$\blacktriangleright${\bf Mental-BERT} \cite{ji2021mentalbertpubliclyavailablepretrained}: A BERT-based model specifically trained on mental health datasets adapted for tasks in this domain.
$\blacktriangleright${\bf BART} \cite{lewis2019bartdenoisingsequencetosequencepretraining}: A powerful encoder-decoder model used for generative tasks in NLP.
$\blacktriangleright${\bf Mental-BART} \cite{yang2023mentalllama}: A variation of BART, fine-tuned on mental health corpora to handle domain-specific challenges.
$\blacktriangleright$\textbf{\citet{yadav-etal-2023-towards} ($SOTA$)}: This baseline implementation focuses on generating text-image feature representations, aligning visual and textual tokens, currently the prior state-of-the-art.

\paragraph{\bf Evaluation Metric.} We present our comparative evaluation using F1 scores (Macro-F1, Weighted-F1, and Micro-F1) on seven classes, demonstrating that our proposed approach outperforms these baselines across the board. To make the results consistent, we ran \citet{yadav-etal-2023-towards} model on \data\ and tabulated scores in the Table \ref{tab:results.tex}.

\begin{table}[!t]
\centering
\resizebox{\columnwidth}{!}{
\begin{tabular}{lcccc}
\toprule
\multirow{2}{*}{Models} & \multicolumn{2}{c}{Depressive Memes} & \multicolumn{2}{c}{Anxiety Memes} \\
\cmidrule(lr){2-3} \cmidrule(lr){4-5}
                        & Macro-F1 & Weigh-F1   & Macro-F1   & Weigh-F1 \\
\midrule
\bf OCR-Text: & & & & \\
\quad BERT & 58.16 & 57.12 & 62.99 & 63.06 \\
\quad Deberta-V3 & 57.99 & 57.40 & 62.03 & 61.86 \\
\quad Mental-BERT & 59.14 & 60.03 & 63.41 & 63.40 \\
\quad BART & 41.56 & 38.35 & 62.82 & 62.62 \\
\quad Mental-BART & 54.19 & 52.55 & 64.97 & 64.88 \\
\midrule
\bf OCR + COMET ($f_k$): & & & & \\
\quad BERT & 58.71 & 58.14 & 61.00 & 61.00 \\
\quad Deberta-V3 & 58.81 & 58.90 & 60.52 & 60.17 \\
\quad Mental-BERT & 60.11 & 60.79 & 63.00 & 63.00 \\
\quad BART & 62.55 & 63.28 & 62.71 & 62.61 \\
\quad Mental-BART & 62.94 & 63.43 & 63.97 & 63.86 \\
\midrule
\bf OCR + LLAVA ($f_k$): & & & & \\
\quad BERT & 63.97 & 66.15 & 62.43 & 62.38 \\
\quad Deberta-V3 & 60.80 & 60.55 & 60.70 & 60.80 \\
\quad Mental-BERT & 65.45 & 63.02 & 63.00 & 63.00 \\
\quad BART & 67.00 & 68.55 & 64.01 & 64.02 \\
\quad Mental-BART & 65.06 & 66.00 & 65.03 & 65.07 \\
\midrule
\bf OCR + GPT-4o ($f_k$): & & & & \\
\quad BERT & 58.65 & 59.05 & 62.01 & 62.02 \\
\quad Deberta-V3 & 59.87 & 59.61 & 61.33 & 62.40 \\
\quad Mental-BERT & 60.80 & 60.29 & 62.38 & 62.52 \\
\quad BART & 63.32 & 64.09 & 64.00 & 64.05 \\
\quad Mental-BART & 63.26 & 64.82 & 65.51 & 65.54 \\
\midrule
\bf \citet{yadav-etal-2023-towards} ($SOTA$): & 63.58 & 64.59 & 64.21 & 65.34 \\
\midrule
\rowcolor{green!24} \bf M3H ($Proposed$): & \textbf{67.52} & \textbf{68.79} & \textbf{70.00} & \textbf{70.10} \\
\midrule
\quad -- (FCS+RAG) & 65.20 & 66.75 & 64.28 & 64.97 \\
\quad -- (OCR+RAG) & 66.06 & 66.01 & 62.54 & 62.53 \\
\quad -- FCS & 58.50 & 58.91 & 62.82 & 62.62 \\
\midrule
$\Delta_{M3H-SOTA}(\%)$ & \textcolor{blue}{$\uparrow 4.94$} & \textcolor{blue}{$\uparrow 4.20$} & \textcolor{blue}{$\uparrow 5.79$} & \textcolor{blue}{$\uparrow 4.66$} \\
\bottomrule
\end{tabular}}
\caption{\model's Performance and Ablation Study. \model\ clearly outperforms all 20 variations of baseline across both datasets: \data\ and \texttt{RESTORE} on both macro- and weighted-F1. It is worth noting that we perform multilabel classification on RESTORE, unlike on \data, which is standard multiclass classification. Due to space constraints, we enclose additional ablations and metrics in Appendix (c.f. Table \ref{tab:ablation}).}
\label{tab:results.tex}
\vspace{-7mm}
\end{table}

\begin{figure}[t]
    \centering
    \includegraphics[width=0.75\columnwidth]{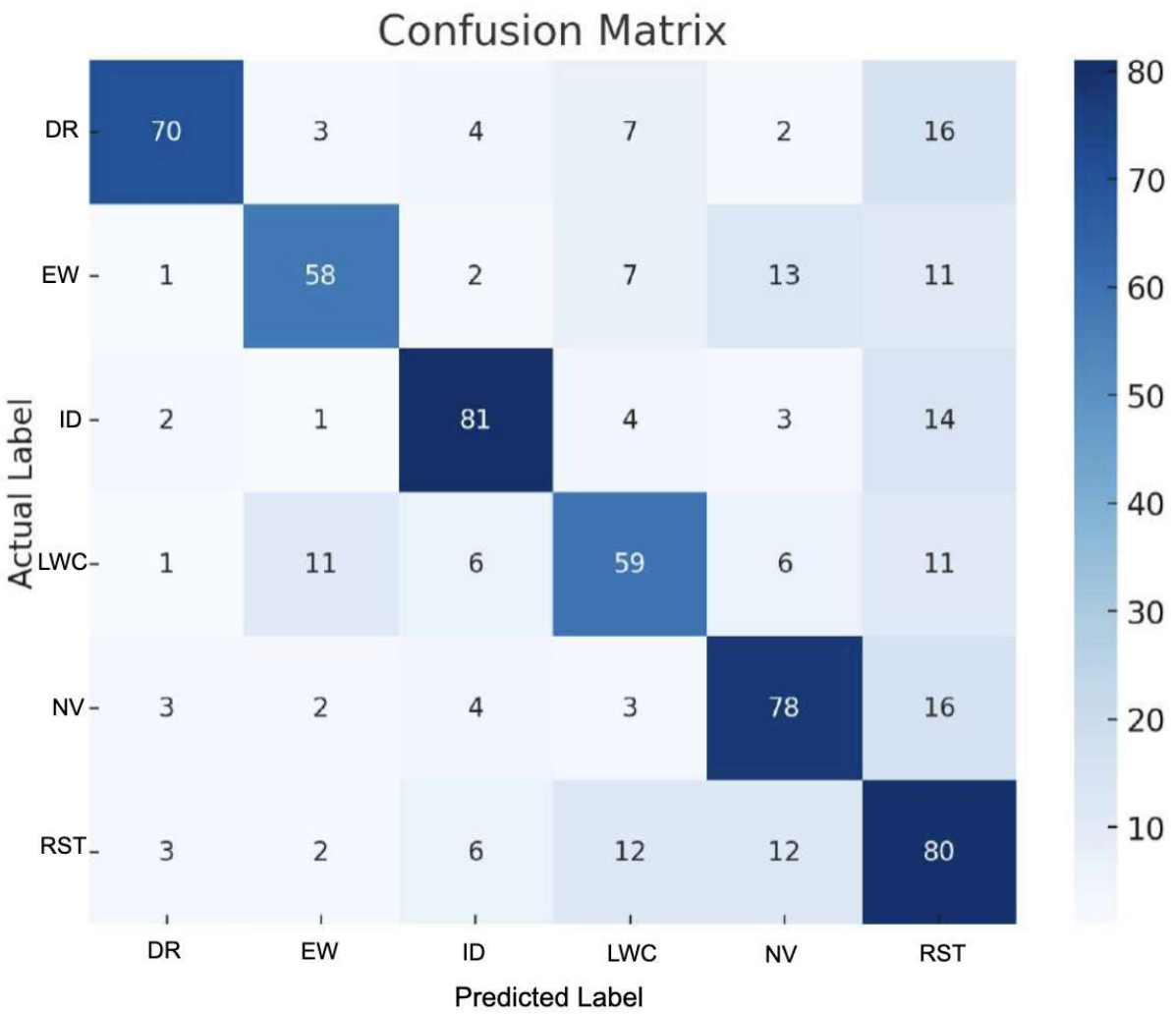}
    \caption{Confusion matrix computed on our proposed dataset, \data\ using our model, \model.}
    \label{fig:confusionmat}
    \vspace{-5mm}
\end{figure}

\begin{figure*}[h]
    \centering
    \includegraphics[width=\textwidth]{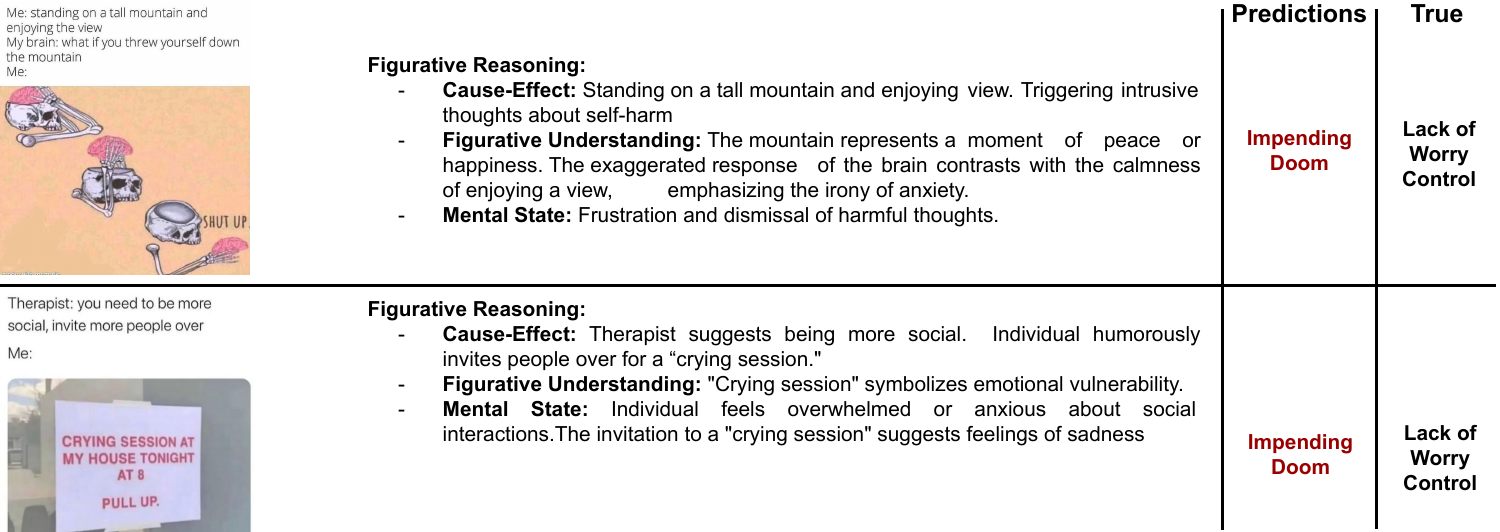}
    \caption{Error Analysis. We attempt to connect the dots between misclassification and figurative reasoning. Evidently, the reasoning module is able to capture underlying nuances; however, we observe label confusion due to which misclassification happens.}
    \label{fig:erroranalysis}
\end{figure*}

\subsection{Performance Comparison}
Our proposed model, \model, shows substantial improvements across 20 variations of baselines, as shown in Table \ref{tab:results.tex}. Primarily, we experiment with standalone OCR text, where Mental-BERT performed best with a score of $59.14\%$ and $60.03\%$ on \data, whereas on \texttt{RESTORE}, it yields $63.41\%$ and $63.40\%$ on macro- and weighted-F1, respectively. Next, we compare the performance of commonsense reasoning in order to extract figurative nuances. We compare COMET, LLaVA, and GPT-4o and observe that GPT-4o performed better across both datasets. On \data, we observe an increment of $+7.06\%$ and $+12.90\%$ with BART and Mental-BART; whereas, on RESTORE, we observe a marginal improvement of $+0.83\%$ and $+1.01\%$ with Mental-BART on macro- and weighted-F1 metrics, respectively. As a result, we fuse \model\ to consider figurative commonsense nuances from \texttt{GPT-4o} and classifier as BART, which provides the best output. With this setting, \model\ achieves F1 scores of $67.52$ ($+494\%$) and $68.79$ ($+4.20\%$), and $70.09$ on macro-, weighted- and micro-F1. On the other hand, we observe a consistent improvement on RESTORE data as well, with $70.00\%$ score on both macro and weighted-F1 score, with an improvement of $+5.79\%$ and $+4.66\%$, respectively.  These results highlight \model’s superiority in classifying symptoms. We present additional metrics and results in the Appendix (c.f. Table \ref{tab:ablation}).

\paragraph{\bf Ablation Study.}  
To better understand the contributions of each component in \model, we conduct an ablation study. We assess the impact of each module on the model's performance, as shown in Table \ref{tab:results.tex}. Our findings indicate a significant decline in \model's performance upon the removal of the figurative reasoning and knowledge module. Specifically, the weighted-F1 score for the depression dataset dropped from $68.79$ to $65.20$ when the FCS-RAG was removed. Additionally, the score further decreased to $66.06$ with the removal of the OCR-RAG. A similar trend is observed in the \data, where macro-F1 score falls from $70.10$ to $64.97$ following the removal of the FCS-RAG and further falls to $62.53$ upon the removal of the OCR-RAG. Notably, removing the FCS module results in scores of $58.91$ and $62.62$ for the \texttt{RESTORE} and \data\ datasets, respectively, showing that both modules contribute positively. We further observe an increment of $+5.09\%$ to quantify the contribution of the RAG module (c.f. Appendix; Figure \ref{fig:ablation}). We present additional ablations in the Appendix due to space constraints (c.f. Section \ref{abalationstudyappendix} and Table \ref{tab:ablation}).

\subsection{Error Analysis}
In this section, we present a two-way error analysis of \model\ performance through both quantitative and qualitative aspects.

\paragraph{\bf Quantitative Analysis.} We report the confusion matrix for our model in \ref{fig:confusionmat}. We observe several pairs with misclassification ($\geq 15\%$). The most significant confusion is between the classes `Lack of Worry Control' and `Impending Doom', where $16\%$ of the former were misclassified as `Impending Doom'. This confusion can be attributed to the similarity in the emotional expression of the two classes, particularly in situations where anxiety involves intrusive thoughts, which is hard to understand casually for even humans. We observe a similar confusion between `Restlessness' vs `Lack of Worry Control', with a confusion rate of $12.2\%$.
Evidently, \model\ tends to confuse the emotional content, where the individual is struggling to manage their emotions, suggesting other labels.

\paragraph{\bf Qualitative Analysis.}
We examine the performance of \model\ and observe that the model sometimes misinterprets the humor or figurative undertones of the text. We present a few such cases in Figure \ref{fig:erroranalysis}. For instance, in case \#1, an individual standing on a tall mountain expresses an intrusive thought about jumping. The correct label is `Lack of Worry Control', reflecting the person's struggle to manage these intrusive thoughts; however, the model predicted `Impending Doom' as a symptom likely due to the high-stakes nature of the situation. 
In case \#2, a meme about a `crying session' is labeled as `Lack of Worry Control', but the model predicts `Restlessness'. The underlying sarcasm, where the individual uses a `crying session' to cope with social anxiety, indicates emotional difficulty rather than physical restlessness; however, \model\ focuses on the anxious behavior implied by the invitation, resulting in misclassification.

\subsection{Human Evaluation}
To assess the quality of the commonsense reasoning generated by GPT-4 for both anxious and depressive memes, we conduct an human evaluation for both datasets. We select a total of 40 random samples, 20 from each dataset. Each meme sample includes figurative reasoning that includes three attributes: {\em cause and effect}, {\em figurative understanding}, and {\em mental state}. We ask human evaluators to rate the commonsense reasoning on the likert scale of 1 (poor) to 5 (excellent). Evidently, as shown in Figure \ref{fig:humaneval}, the majority of feedback in both datasets falls between $4$ and $5$, indicating that the figurative-cum-commonsense reasoning is generally well-perceived, aligning with human interpretations of the underlying sarcasm and mental health context.

\section{Discussion, Ethics, and Limitations}
Our findings emphasize the importance of integrating visual and textual commonsense reasoning to accurately classify depressive and anxious symptoms in memes. While M3H demonstrates strong performance, there is room for improvement. Relying solely on online memes limits the framework’s ability to capture complex, trendy nuances due to a lack of diverse knowledge. Although LLMs enhance performance through commonsense reasoning, they increase computational complexity and may produce responses misaligned with human reasoning or fail to capture meme-specific nuances, which are critical in mental health contexts. Importantly, our framework is not intended to diagnose mental health disorders but to analyze social media posting behavior related to depressive and anxious meme content. Future work will focus on improving the diversity, factuality, and faithfulness of generated reasoning. Expanding our approach to include other domains of figurative communication, such as political cartoons or digital art, could provide a broader understanding of memes and their contextual issues.

\section{Conclusion}
Social media has become a prominent platform for expressing mental health issues through memes. In this work, we present \model, a novel framework that leverages GPT-4 for figurative reasoning and a BART-based transformer for identifying symptoms depicted in memes. Through rigorous benchmarking against six competitive baselines (20 variations), including state-of-the-art models, we demonstrate \model's superiority on both the newly proposed \data\ dataset for anxious symptom classification and the \texttt{RESTORE} dataset for depressive symptom classification. Using macro-F1 and weighted-F1 scores, \model achieves significant improvements of $4.94\%$ and $5.79\%$ in macro-F1, and $4.20\%$ and $4.66\%$ in weighted-F1 across both datasets. Our study is supported by exhaustive human evaluation, qualitative and quantitative analyses, and a discussion on ethics and limitations.

\section*{Acknowledgment}
The authors acknowledge the support of the Infosys Foundation through CAI at IIIT-Delhi and the iHub-Anubhuti-iiitd Foundation.

\bibliographystyle{ACM-Reference-Format}
\bibliography{main}

\clearpage
\appendix

\section{Data Collection}\label{dataCollection}
We gather data from various online sources, including Reddit and Instagram, which are popular platforms known for unfiltered expression and meme sharing. The anonymity provided by these platforms allows for more expressive meme content. To collect data from Reddit, we use Reddit's API and developed a scraper to extract images from specific subreddits, setting a maximum limit of 3,000 images per subreddit. For Instagram, we applied a similar approach to extract images from public profiles using our scraper. The subreddits, such as \textit{r/2meirl4meirl, r/anxietymemes, r/healthanxietyhumor, r/sa\_memetherapy, r/socialanxietymemes}, and \textit{r/TrollCoping}, were selected based on Google search results identifying popular mental health subreddits with frequent meme-posting about mental health struggles. These subreddits capture diverse anxiety-related expressions, including depression-related content (e.g., impending doom and restlessness from sleep disturbances in \textit{r/2meirl4meirl}) and raw, unfiltered memes reflecting anxious characteristics like dissociation and distorted reality (e.g., \textit{r/bpd\_memes}). For Instagram, we selected profiles like \textit{anxietian, anxietytwitter}, and \textit{anxietymemetherapy}, using the WorryWords lexicon~\cite{mohammad2024worrywords}, a comprehensive list of over 44,000 English words scored for their association with anxiety. Profiles were chosen based on the presence of anxiety-related keywords in their usernames, bios, or handles. The collected memes were annotated with their expressed mental health symptoms based on a predefined set of symptoms. This approach ensures a broad spectrum of anxiety-related expressions, enriched by contributions from individuals with varied backgrounds and mental health conditions.

\section{Experimental Setup} \label{experimentalsetup}

For the baseline's training, we have inferred all the models from Hugging Face and trained for 10 epochs. It took approximately 50 minutes for BERT-based models and 1.2 hours for BART-based models, using 3 NVIDIA Tesla V100 GPUs. For the proposed model, we have trained for 10 epochs, with each epoch taking about 30 minutes. The inference time for classifying a depression category is approximately 5 seconds. All training and inference operations were conducted on 3 NVIDIA Tesla V100 GPUs. 

We employed the AdamW optimizer with a learning rate of $5 × 10^-5$, beta1 of 0.9, and beta2 of 0.999. To prevent overfitting, we applied a dropout rate of 0.2 and a weight decay of $1 × 10^-2$. For single-label classification tasks, we used Cross Entropy Loss, while Binary Cross Entropy Loss was employed for multi-label scenarios. The model's performance was evaluated using Micro, Macro, and Weighted F1 Scores to provide a comprehensive assessment across different class distributions and task types.

The learning rate scheduler was set to constant, and we used a batch size of 16. The RMS Norm Epsilon was set to $1 × 10^-5$, and the Adam Epsilon to $1 × 10^-8$. The maximum sequence length was capped at 512 tokens. Gradient clipping was applied with a threshold of 1.0 to stabilize training.

These hyperparameters were carefully tuned to optimize model performance while balancing computational efficiency. The detailed hyperparameters are provided in Table 1, which includes additional parameters such as warmup steps and specific optimizer settings.

This configuration allowed us to effectively train and evaluate our models across various mental health meme classification tasks, accounting for both single-label and multi-label scenarios, while ensuring robust performance metrics through the use of diverse F1 score calculations.

\begin{table}[h]
\centering
\begin{tabular}{ll}
\hline
\centering
\textbf{Hyperparameter} & \textbf{Value} \\
\hline
Learning Rate (lr)& $5 \times 10^{-5}$ \\
Adam Beta1              & 0.9                 \\
Adam Beta2              & 0.999               \\
Adam Epsilon            & $1 \times 10^{-8}$  \\
RMS Norm Epsilon        & $1 \times 10^{-5}$  \\
Dropout                 & 0.2                 \\
Batch Size              & 16                  \\
Learning rate schedular              & constant                  \\
\hline
Lora Rank (r) & 16 \\
Lora alpha ($\alpha$) & 8 \\
Target modules & $W_q, W_k, W_v, W_o$ \\
\hline
\end{tabular}
\caption{Hyperparameters used in the experiment.}
\label{tab:hyperparameters}
\end{table}

\begin{figure}[b]
    \centering
    \includegraphics[width=\columnwidth]{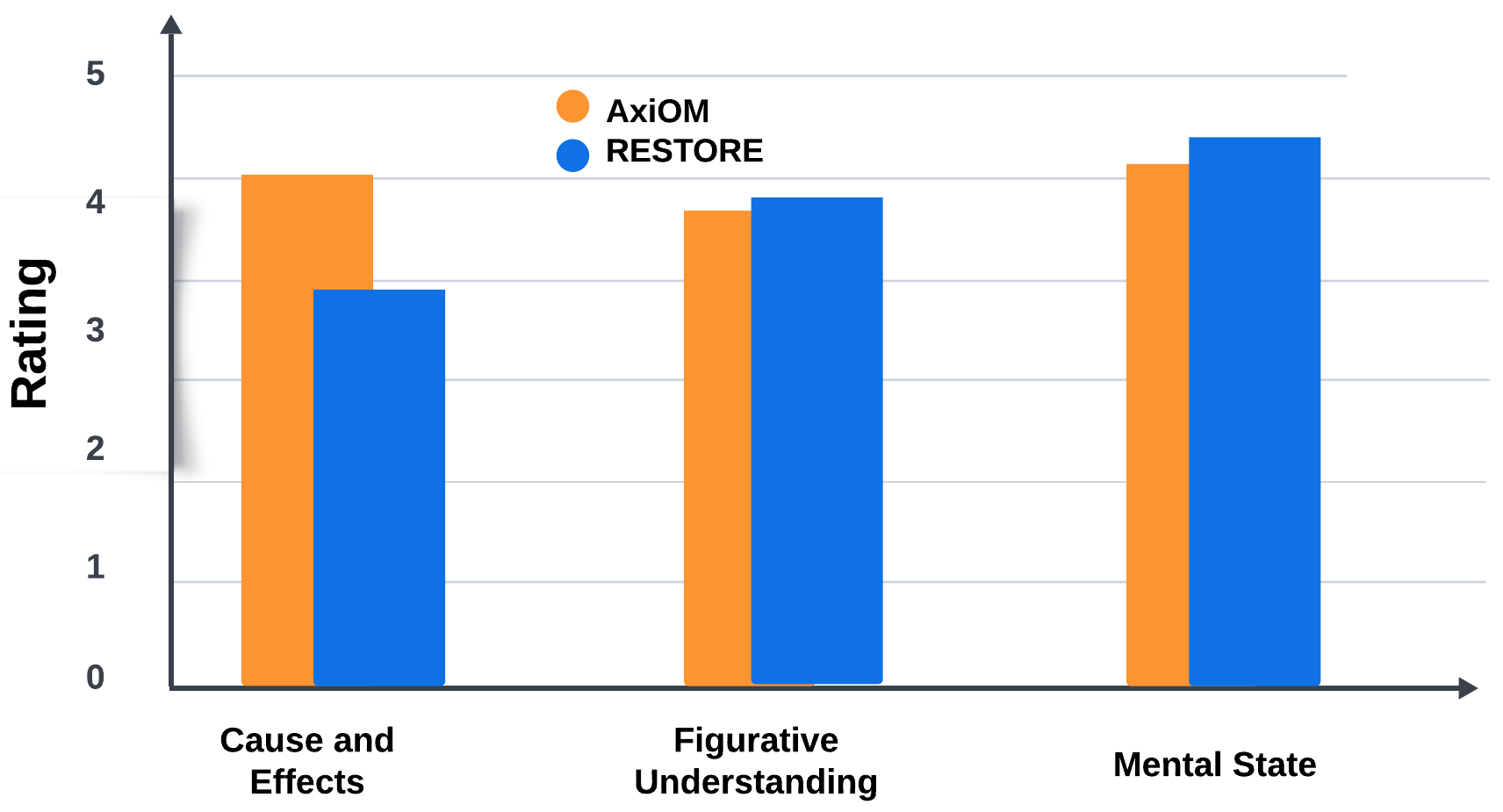}
    \caption{Human Evaluation for commonsense reasoning using GPT-4o for each figurative reasoning attribute on both dataset: \data\ and \texttt{RESTORE}.}
    \label{fig:humaneval}
\end{figure}

\begin{table*}[t]
\centering
\resizebox{\textwidth}{!}{
\begin{tabular}{ccccccccccc}
\toprule
\multirow{2}{*}{\bf OCR} & \multirow{2}{*}{\bf OCR-RAG} & \multirow{2}{*}{\bf FCS} & \multirow{2}{*}{\bf FCS-RAG} & \multicolumn{3}{c}{\bf Depression} & & \multicolumn{3}{c}{\bf Anxiety} \\ \cmidrule{5-7} \cmidrule{9-11}
& & & & \bf Macro-F1 & \bf Weighted-F1 & \bf Micro-F1 & & \bf Macro-F1 & \bf Weighted-F1 & \bf Micro-F1 \\
\midrule
\text{\ding{51}} & \text{\ding{55}} & \text{\ding{55}} & \text{\ding{55}} & 58.50 & 58.91 & 62.29 & & 62.82 & 62.62 & 62.21\\
\text{\ding{51}} & \text{\ding{51}} & \text{\ding{55}} & \text{\ding{55}} & 66.01 & 67.33 & 68.60 & & 62.54 & 62.53 & 62.05 \\
\midrule
\text{\ding{55}} & \text{\ding{55}} & \text{\ding{51}} & \text{\ding{55}} & 61.18 & 62.05 & 66.84 & & 57.00 & 58.01 & 56.84 \\
\text{\ding{55}} & \text{\ding{55}} & \text{\ding{51}} & \text{\ding{51}} & 66.42 & 67.65 & 69.24 & & 60.45 & 60.69 & 60.46\\
\midrule
\text{\ding{51}} & \text{\ding{55}} & \text{\ding{51}} & \text{\ding{55}} & 63.32 & 64.09 & 66.22 & & 64.28 & 64.97 & 64.20 \\
\text{\ding{51}} & \text{\ding{55}} & \text{\ding{51}} & \text{\ding{51}} & 65.20 & 66.75 & 69.01 & & 64.94 & 64.95 & 64.84\\
\text{\ding{51}} & \text{\ding{51}} & \text{\ding{51}} & \text{\ding{55}} & 66.06 & 66.01 & 67.92 & & 61.82 & 61.79 & 62.21 \\
\text{\ding{51}} & \text{\ding{51}} & \text{\ding{51}} & \text{\ding{51}} & \textbf{67.52} & \textbf{68.79} & \textbf{70.09} & & \textbf{70.00} & \textbf{70.00} & \textbf{69.38} \\

\bottomrule
\end{tabular}}
\caption{Experimental results for various model configurations}
\label{tab:ablation}
\end{table*}

\section{Additional Ablation Study}
\label{abalationstudyappendix}
\begin{figure}[b]
    \centering
    \includegraphics[width=\columnwidth]{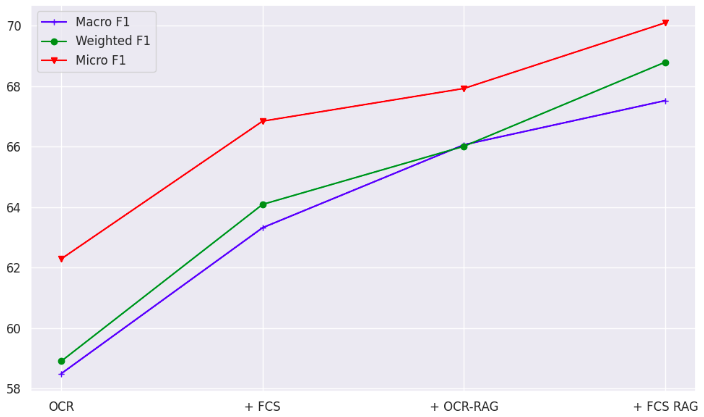}
    \caption{There is a significant increase in the score when RAG is added into the system $\Delta$ RAG = 5.40 \% on Depressive meme dataset.}
    \label{fig:ablation}
\end{figure}

Each component of the \model\ module plays a crucial role in enhancing performance, as demonstrated in Table \ref{tab:ablation}. The table outlines how removing one or more components affects the overall model performance. Here, OCR refers to Optical Character Recognition of text from the meme, while \textbf{FCS} (Figurative Common Sense) represents deep understanding extracted from multimodal large language models. We report \textbf{Macro F1}, \textbf{Weighted F1}, and \textbf{Micro F1} scores on both the \textbf{RESTORE} and \textbf{AxioM} datasets. From the ablation results, we observe that incorporating all components yields the highest scores across all metrics. On the other hand, in Figure \ref{fig:ablation}, the performance improvement with each additional component is visualized. Notably, the system exhibits a significant performance boost of $\Delta$RAG = 5.40 after introducing knowledge-fusion, indicating its effectiveness in enhancing the system's overall capability.

\section{Prompts and Examples} 
\label{promptsandexamples}
\begin{tcolorbox}[width=\columnwidth,title=,colupper=black]\small
Analyze the following anxiety meme image to extract common sense reasoning in the form of triples.. 
    These relationships should capture the following elements:
\begin{itemize}
    \item 1. Cause-Effect: Identify concrete causes or results of the situation depicted in the meme.

    \item 2. Figurative Understanding: Capture underlying metaphors, analogies, or symbolic meanings that convey the meme's deeper message, including any ironic or humorous undertones.

    \item 3. Mental State: Capture specific mental or emotional states depicted in the meme.
\end{itemize}
\end{tcolorbox}

\begin{tcolorbox}[width=\columnwidth,title=,colupper=black]\small
Analyze the following depression meme image to extract common sense reasoning in the form of triples.
    These relationships should capture the following elements:
\begin{itemize}
    \item 1. Cause-Effect: Identify concrete causes or results of the situation depicted in the meme.

    \item 2. Figurative Understanding: Capture underlying metaphors, analogies, or symbolic meanings that convey the meme's deeper message, including any ironic or humorous undertones.

    \item 3. Mental State: Capture specific mental or emotional states depicted in the meme.
\end{itemize}
\end{tcolorbox}
\begin{table*}
\centering
\footnotesize
\resizebox{\textwidth}{!}{%
\begin{tabular}{p{58em}} 
\toprule
\textbf{System prompt:} You specialize in analyzing mental health behaviors through social media posts. Your task is to classify the mental health issue depicted in a person's post from the following categories: ['Difficulty Relaxing', 'Excessive Worry', 'Impending Doom', 'Lack of Worry Control', 'Nervousness', 'Restlessness']. \\
\midrule
\textbf{Example-1:} \\
<|ocr\_text|> "I DONT ALWAYS SLEEP AT NIGHT BUT WHEN I DO, I DONT" \\
<|commonsense figurative explanation|> Here are some common sense reasoning triples extracted from the anxiety meme: \\
Cause-Effect \\
\quad 1. Cause: Not sleeping regularly at night \\
   \quad \quad Effect: Feeling tired or anxious during the day. \\
\quad 2. Cause: Drinking alcohol (implied with the beer bottle) \\
   \quad \quad Effect: Difficulty in sleeping or restlessness at night. \\
Figurative Understanding \\
\quad 1. Metaphor: "I don’t always sleep at night" \\
   \quad \quad Interpretation: Suggests a chaotic or irregular sleep pattern that many can relate to, often due to anxiety. \\
\quad 2. Irony: "But when I do, I don’t" \\
   \quad \quad Interpretation: Implies that even when he does sleep, it's not restful, highlighting the contradiction in seeking rest yet feeling unrested. \\
Mental State \\
\quad 1. Mental State: Anxiety or restlessness at night \\
   \quad \quad Description: Reflects the inner turmoil of being unable to sleep due to worries or stress. \\
\quad 2. Mental State: Humor or sarcasm \\
   \quad \quad Description: The playful tone of the statement conveys a sense of humor about the struggles of insomnia or disrupted sleep patterns. \\
<|Assistant|> The mental health disorder of the person for this post is: \textbf{Difficulty Relaxing} \\
\midrule
\textbf{Example-2:} \\
<|ocr\_text|> "am I exhausted? Yes will I go to bed earlier? No" \\
<|commonsense figurative explanation|> Here are some triples that capture the common sense reasoning from the anxiety meme image: \\
Cause-Effect: \\
\quad 1. Cause: The individual feels exhausted. \\
\quad 2. Effect: Despite this exhaustion, the individual will not go to bed earlier. \\
Figurative Understanding: \\
\quad 1. Symbolic Meaning: The exhaustion represents a common struggle with prioritizing self-care. \\
\quad 2. Humorous Undertone: The light-hearted refusal to go to bed earlier despite acknowledgment of exhaustion illustrates a relatable, ironic attitude towards personal habits. \\
Mental State: \\
\quad 1. Mental State: The individual is feeling exhausted. \\
\quad 2. Emotional State: There is a sense of defiance or acceptance of unhealthy habits, possibly mixed with humor or resignation. \\
<|Assistant|> The mental health disorder of the person for this post is: \textbf{Difficulty Relaxing} \\
\midrule
\textbf{Your turn:} \\
<|ocr\_text|> "Dad: you're up early" - "Me:" - "Dad: you never went to sleep did you" \\
<|commonsense figurative explanation|> Here are the extracted triples based on the anxiety meme: \\
Cause-Effect \\
\quad 1. Cause: The child is up early. \\
   \quad \quad Effect: The dad questions the child's sleep pattern. \\
\quad 2. Cause The child did not go to sleep. \\
   \quad \quad Effect: The dad realizes the child may be sleep-deprived or acting unusually. \\
Figurative Understanding: \\
\quad 1. Figurative Meaning The dad's smile symbolizes an awareness of common teenage behavior (e.g., staying up late). \\
   \quad 2. Underlying Message: It humorously highlights a typical parent-child dynamic regarding sleep schedules. \\
\quad 3. Irony The child being up early might typically be seen as positive, but it signifies a lack of sleep instead, turning a seemingly good situation into a concerning one. \\
Mental State: \\
\quad 1. Mental State of the Dad: Amusement or bemusement at the child's situation. \\
   \quad \quad Emotion: Curiosity mixed with a hint of concern. \\
\quad 2. Mental State of the Child Possibly anxious or guilty about not sleeping. \\
   \quad \quad Emotion: Discomfort or embarrassment due to the revelation of staying awake all night. \\
<|Assistant|> The mental health disorder of the person for this post is: \textcolor{red}{\textbf{||}} M3H Prediction: \textbf{Difficulty Relaxing}\\
\bottomrule
\end{tabular}
}
\caption{This table presents the detailed end-to-end prompt used for our proposed model, \model, along with the corresponding prediction.}
\label{promptInputExample}
\end{table*}

\end{document}